\pdfoutput=1

\documentclass[11pt]{article}

\usepackage{ACL2023}

\usepackage{times}
\usepackage{latexsym}

\usepackage[T1]{fontenc}

\usepackage[utf8]{inputenc}

\usepackage{microtype}

\usepackage{graphicx}
\usepackage{multicol}
\usepackage{multirow}
\usepackage{amsmath}
\usepackage{amssymb}
\usepackage{bbding}
\usepackage{subfigure}
\usepackage{colortbl}
\usepackage{pifont}
\usepackage{bbm}
\usepackage{makecell}
\usepackage{bbding}
\usepackage{mathrsfs}
\usepackage{bm}
\usepackage{fontawesome}
\usepackage{booktabs}
\usepackage{tabu}
\usepackage{xcolor}
\usepackage{makecell}
\usepackage{colortbl}
\usepackage{xcolor}

\definecolor{Gray}{gray}{0.93}

\newcommand{\benchmark}{CodeArena}
\newcommand{\instruct}{SynCode-Instruct}
\newcommand{\baseline}{Qwen2.5-SynCoder}

\definecolor{dkgreen}{rgb}{0,0.6,0}
\definecolor{gray}{rgb}{0.5,0.5,0.5}
\definecolor{mauve}{rgb}{0.58,0,0.82}

\title{Evaluating and Aligning CodeLLMs on Human Preference}

\author{
  Jian Yang\textsuperscript{\rm 1},
  Jiaxi Yang\textsuperscript{\rm 2,3},
  Ke Jin, 
  Yibo Miao\textsuperscript{\rm 4},  
  {\bf Lei Zhang}\textsuperscript{\rm 2,3},  
  {\bf Liqun Yang}, \\
  {\bf Zeyu Cui}\textsuperscript{\rm 1},   
  {\bf Yichang Zhang}\textsuperscript{\rm 1},  
  {\bf Binyuan Hui}\textsuperscript{\rm 1},
  {\bf Junyang Lin}\textsuperscript{\rm 1} \\
  \textsuperscript{\rm 1}Alibaba Group; \textsuperscript{\rm 2}Shenzhen Institute of Advanced Technology, Chinese Academy of Sciences;\\
  \textsuperscript{\rm 3}University of Chinese Academy of Sciences; \textsuperscript{\rm 4}Shanghai Jiao Tong University \\
  \texttt{\{yj411294,binyuan.hby,junyang.ljy\}@alibaba-inc.com} \\
}

\begin{document}

\maketitle

\begin{abstract}
Code large language models (codeLLMs) have made significant strides in code generation. Most previous code-related benchmarks, which consist of various programming exercises along with the corresponding test cases, are used as a common measure to evaluate the performance and capabilities of code LLMs. However, the current code LLMs focus on synthesizing the correct code snippet, ignoring the alignment with human preferences, where the query should be sampled from the practical application scenarios and the model-generated responses should satisfy the human preference. To bridge the gap between the model-generated response and human preference, we present a rigorous human-curated benchmark \textbf{\benchmark{}} to emulate the complexity and diversity of real-world coding tasks, where 397 high-quality samples spanning 40 categories and 44 programming languages, carefully curated from user queries. Further, we propose a diverse synthetic instruction corpus \instruct{} (nearly 20B tokens) by scaling instructions from the website to verify the effectiveness of the large-scale synthetic instruction fine-tuning, where \baseline{} totally trained on synthetic instruction data can achieve top-tier performance of open-source code LLMs.
The results find performance differences between execution-based benchmarks and \benchmark{}. Our systematic experiments of \benchmark{} on 40+ LLMs reveal a notable performance gap between open SOTA code LLMs (e.g. Qwen2.5-Coder) and proprietary LLMs (e.g., OpenAI o1), underscoring the importance of the human preference alignment.\footnote{\url{https://codearenaeval.github.io/ }}

\end{abstract}

\section{Introduction}
\label{sec:introduction}

Advanced large language models (LLMs)\cite{gpt4,claude} have demonstrated impressive performance across a wide range of tasks, particularly excelling in code completion and generation. Code capabilities have established LLMs as essential productivity tools in software engineering. 
Recently, open code-specific LLMs, such as StarCoder\cite{starcoder}, DeepSeekCoder~\cite{deepseek_coder}, and QwenCoder~\cite{qwen25coder}, have made significant progress, achieving performance on fundamental code generation tasks~\cite{mbpp,multiple} that approaches the level of top-tier proprietary models. Moreover, their open and transparent model weights address developers' concerns about privacy, enabling the deployment of localized code assistants.


\begin{figure}[t]
\centering
\includegraphics[width=1.0\linewidth]{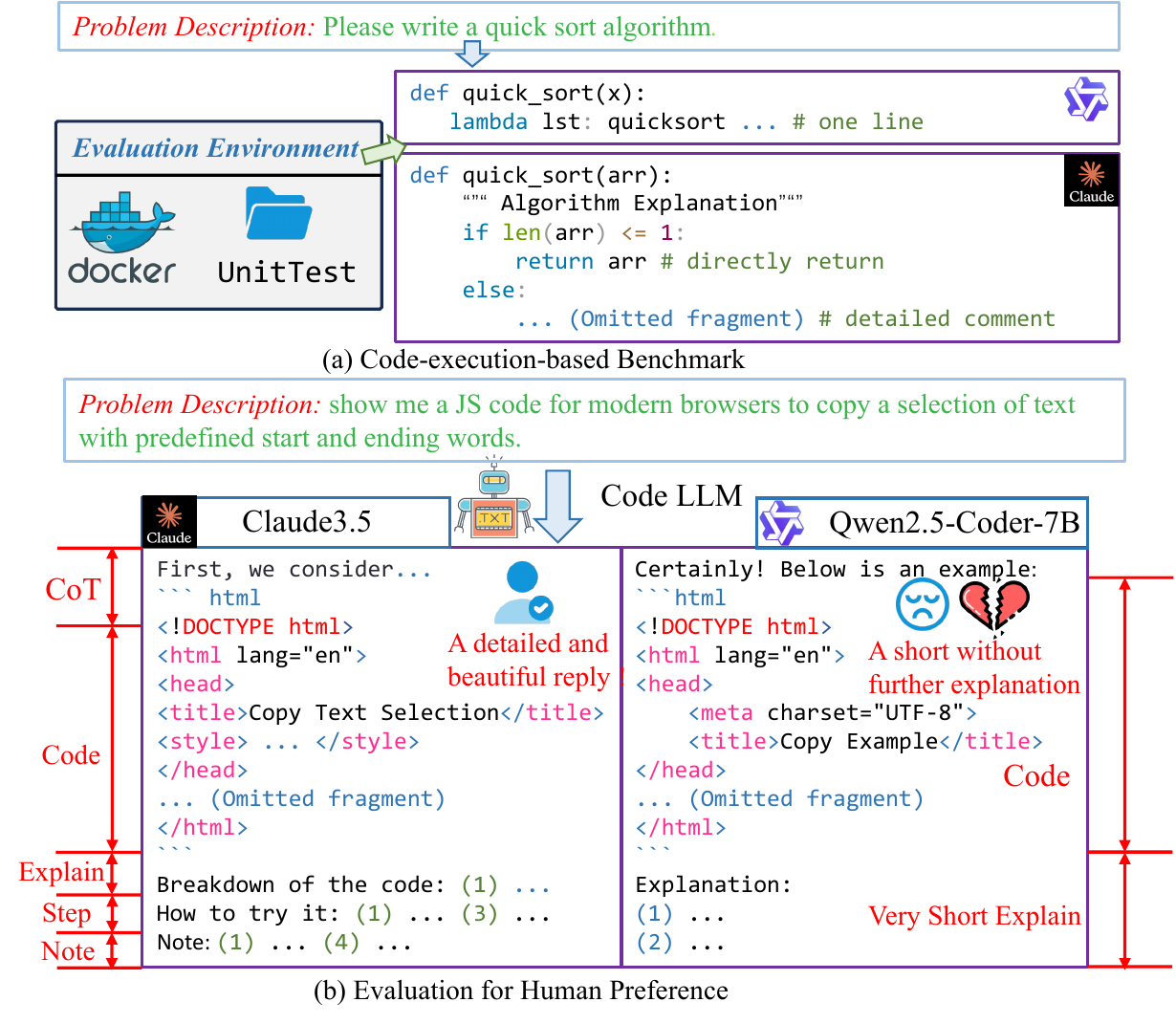}
\caption{A comparison between the GPT4o with better human preference and Qwen2.5-Coder-7B-Instruct. Qwen2.5-Coder-7B-Instruct solves the user question by simply replying with the code snippet without details.}
\vspace{-18pt}
\label{fig:intro}
\end{figure}

With the advancing code capabilities of LLMs, effectively evaluating performance on code-related tasks has emerged as a challenge. Popular code-related benchmarks typically focus on self-contained function snippets, relying on a limited number of test cases to verify code correctness, such as HumanEval~\cite{codex}, MBPP~\cite{mbpp} and BigCodeBench~\cite{zhuo2024bigcodebench}. While recent efforts have expanded the scope of test cases~\cite{evalplus}, tasks~\cite{ds1000} and programming languages~\cite{mceval,naturalquestions}, these benchmarks remain constrained to validating the correctness of generated code snippets. 
However, ChatBot Arena~\citep{chatbot_arena} has demonstrated that alignment between model-generated responses and user preferences is also a critical evaluation criterion. As shown in Figure \ref{fig:intro}, Qwen2.5-Coder primarily generates alone code snippets, while Claude3.5 produces responses that include detailed explanations, well-structured formatting, and code comments, making it more favorable in terms of human preference.
Therefore, there is an urgent need to establish a human preference benchmark specifically for code-related tasks, enabling the community to evaluate and track the alignment between human preferences and model-generated responses in real-world scenarios.
Furthermore, effective data for improving the human preference alignment of codeLLMs remains scarce. Achieving robust alignment across diverse coding tasks poses significant challenges, particularly in terms of the quantity and quality of data required during the supervised fine-tuning (SFT) stage.


To this end, we first introduce a comprehensive human-curated benchmark, \textbf{\benchmark{}}, comprising 397 high-quality samples across 40 categories derived from real-world user queries. Additionally, we develop a diverse synthetic instruction corpus, \textbf{\instruct{}}, containing nearly 20 billion tokens, by scaling instructions from web sources. 
Our extensive evaluation of over nearly 40 large language models (LLMs) using \benchmark{} reveals significant performance differences between code-execution-based benchmarks and our human-curated benchmark. Notably, we observe a substantial performance gap between open-source code LLMs (such as Qwen-Coder) and closed-source LLMs (like the o1 and Claude series), emphasizing the critical role of aligning AI models with human preferences in coding tasks.

The contributions are summarized as follows: (1) We propose \benchmark{} comprised of 397 manually annotated samples, a comprehensive code evaluation benchmark for evaluating the alignment between the model-generated response and human preference, encompassing 7 major categories and 40 subcategories. 
(2) We introduce \instruct{}, the large-scale synthetic code instruction corpora from the website. Based on \instruct{}, an effective coder \baseline{} is used as a strong baseline for \benchmark{}.
(3) We systematically evaluate 40+ LLMs on \benchmark{} and create a leaderboard to dynamically update the results. Notably, extensive experiments suggest that \benchmark{} can effectively measure the alignment between the model-generated response and human preference.
\section{\benchmark{}}

\begin{figure}[h!]
\centering
\includegraphics[width=0.85\columnwidth]{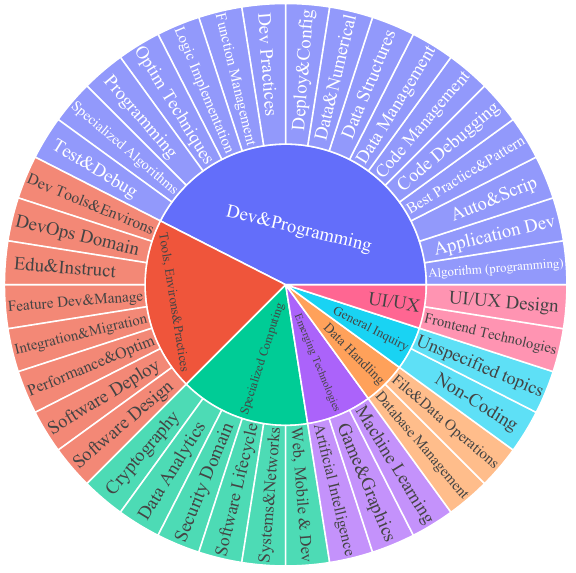}
\caption{Task types of \benchmark{}.}
\vspace{-10pt}
\label{fig:sample_classification}
\end{figure}

\paragraph{Dataset Statistics}
As shown in Figure~\ref{fig:sample_classification} and Table~\ref{tab:detail_data}, \benchmark{} consists of nearly 400 problems. All samples can be classified into 7 main classes and 40 subclasses. Each sample in \benchmark{} includes (\textit{question, gpt-4o-2024-05-13 response, gpt-4o-2024-08-06 response, gpt-4-turbo-2024-04-09 response}) and we adopt the \textit{gpt-4-turbo-2024-04-09} as the baseline in this paper. 
We tokenized the question prompts using the Qwen2.5-Coder tokenizer, resulting in question lengths ranging from 5 to 6736 tokens, with an average length of 291 tokens, as detailed in Table~\ref{tab:detail_data}.

\begin{table}[h!]
\centering
\resizebox{0.85\columnwidth}{!}{
\begin{tabular}{lr}
\toprule
\textbf{Statistics}            & \textbf{Number}       \\
\midrule
\textbf{Problems}            & $397$          \\
User Interface\&Experience            & $45$                 \\
Development\&Programming              & $131$    \\
Specialized Computing                 & $91$                 \\
Tools, Environments, and Application  & $39$    \\
Miscellaneous and General Inquiry     & $62$    \\
Databases\&Data Handling              & $22$    \\
Miscellaneous and General Inquiry     & $7$    \\
\midrule
\textbf{\#Difficulty Level } &                          \\
- Easy/Medium/Hard               & $97$/$173$/$132$ \\
\midrule
\textbf{Length}                                         \\
Question \\
~~~~- \textit{maximum length}   & $6736$ tokens          \\
~~~~- \textit{minimum length}   & $5$ tokens            \\
~~~~- \textit{avg length}       & $291$ tokens         \\
Baseline Answer  \\
~~~~- \textit{maximum length}   & $5913$ tokens          \\
~~~~- \textit{minimum length}   & $7$ tokens            \\
~~~~- \textit{avg length}       & $4517$ tokens         \\
\bottomrule
\end{tabular}}
\caption{\benchmark{} dataset statistics. }
\vspace{-10pt}
\label{tab:detail_data}
\end{table}

\begin{figure*}[h!]
\begin{center}
    \includegraphics[width=1.0\textwidth]{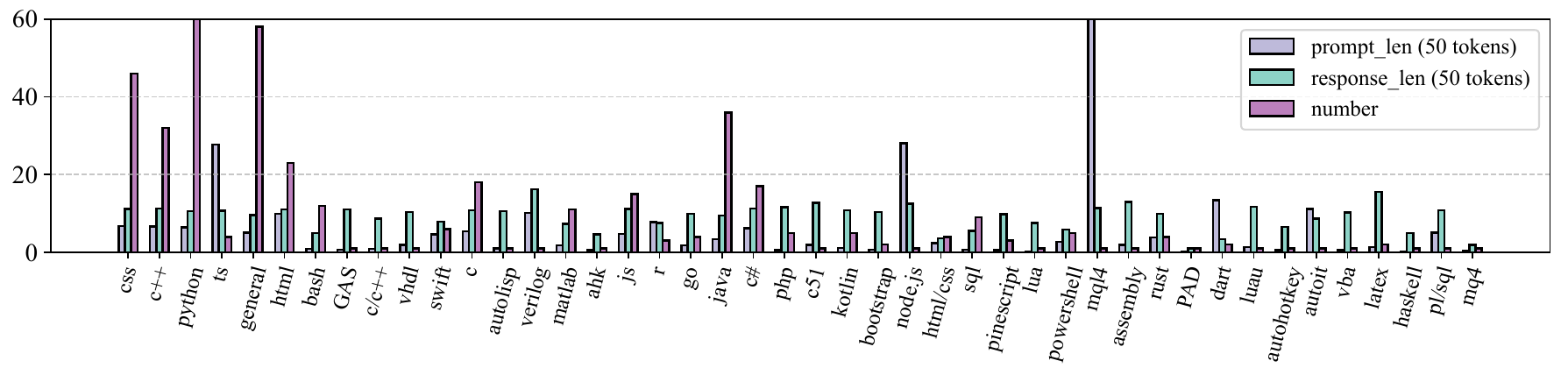}
    \caption{Statistics of programming languages in \benchmark{}. }
    \label{fig:data_statistics}
    \vspace{-15pt}
\end{center}
\end{figure*}
\begin{figure*}[h!]
\begin{center}
    \includegraphics[width=1.0\textwidth]{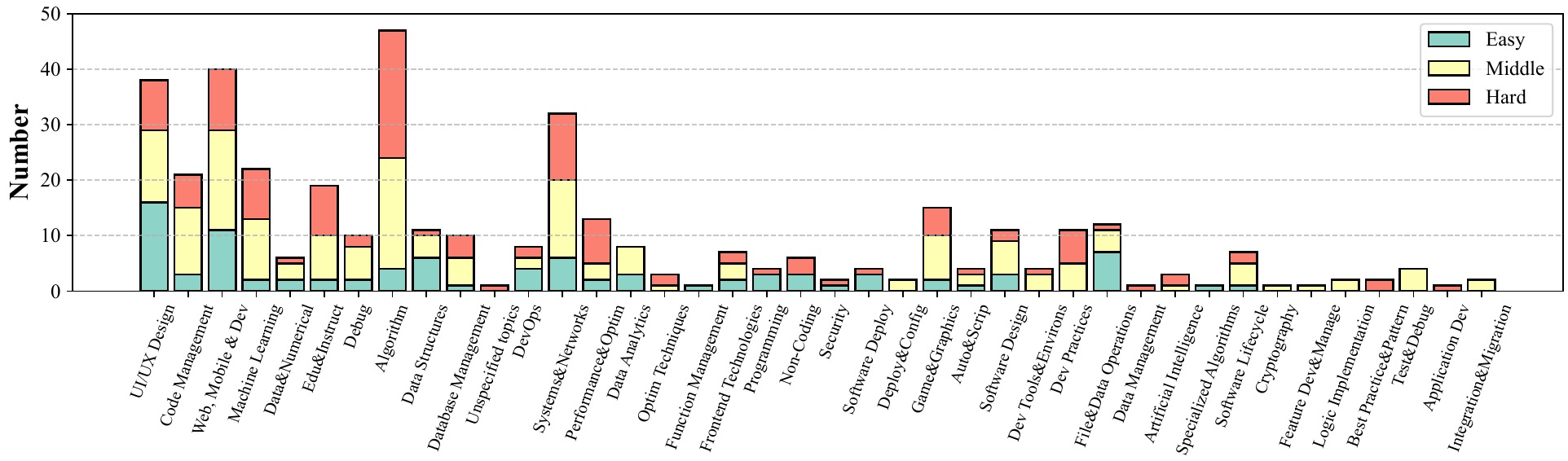}
    \vspace{-5pt}
    \caption{Number of samples of different difficulties (Easy/Medium/Hard) across categories in \benchmark{}. }
    \label{fig:difficulty_statistics}
    \vspace{-10pt}
\end{center}
\end{figure*}

\paragraph{Multiple Programming Languages} Figure~\ref{fig:data_statistics} plots the distribution of programming languages, where we strive to cover common programming languages in \benchmark{}. 
Unlike previous studies~\cite{multiple}, our benchmarks emphasize a diverse range of programming languages that are commonly used in everyday programming tasks. For instance, we have incorporated languages like ``Google Apps Script (GAS)'' and ``PowerShell'' in \benchmark{} to better address the needs of practical Q\&A scenarios.

\paragraph{Difficulty levels of \benchmark{}} Figure \ref{fig:difficulty_statistics} illustrates the difficulty levels of \benchmark{}, where all samples are classified into \textit{easy}, \textit{medium}, and \textit{hard}. 
The majority of the samples are recognized as medium or hard, presenting a significant challenge to LLMs.

\paragraph{Human Annotation \& Quality Control}
To make \benchmark{} a comprehensive evaluation benchmark, we implement a rigorous human annotation process involving 4 full-time employees proficient in various programming languages for human annotation and 4 other senior programming developers for quality check. 
All annotators participate in a annotation tutorial and learn the annotation guidelines. The annotation process involved creating a new question based on the given question, checking the difficulty level (easy/medium/hard) based on the complexity of the prompt, and annotating the corresponding programming languages. Following the classification in Figure \ref{fig:sample_classification}, we uniformly sample 2K samples and assign them to annotators. The annotators select 822 suitable original samples to create queries. The process includes regular quality checks and feedback sessions to maintain high standards throughout the annotation phase, which results in a diverse and well-curated dataset spanning multiple programming languages and tasks, suitable for evaluating and improving alignment between the human preference and model-generated response. The other four senior programming developers vote on the same issue to determine whether it is valid and can be resolved. Finally, 397 samples are kept (at least 3 checkers reach a consensus) to from \benchmark{}, considering the cost of the LLM-as-a-judge.

\begin{table*}[h!]
\centering
\resizebox{1.0 \textwidth}{!}{
\begin{tabular}{l|cccccc}
\toprule
\bf Benchmark& \bf \makecell[c]{\#Programming\\ Languages} & \bf \#Task & \bf Source & \bf \#Languages & \bf Evaluation & \bf \makecell[c]{Human \\ Annotation} \\
\midrule
HumanEval~\citep{codex}& 1& 1& Human Creation& 1 & Execution& \textcolor{green}{\ding{51}} \\
MBPP~\citep{mbpp}& 1& 1& Human Creation& 1 & Execution& \textcolor{green}{\ding{51}} \\
LiveCodeBench~\citep{livecodebench}& 1& 4& Scraped from Code Contest Website& 1 & Execution& \textcolor{green}{\ding{51}} \\
MultiPl-E~\citep{multiple}& 24& 1& Translated from HumanEval \& MBPP& 1 & Execution& \textcolor{red}{\ding{55}}\\
McEval~\citep{mceval}& 40& 3& Human Creation& 1 & Execution& \textcolor{green}{\ding{51}} \\
MdEval~\citep{mdevl}& 18& 3& Human Creation& 1 & Execution& \textcolor{green}{\ding{51}} \\
CruxEval~\citep{cruxeval}& 1& 2& LLM Generation& 1 & Execution& \textcolor{red}{\ding{55}}\\
NaturalCodeBench~\citep{ncb}& 2& 6& Scrape \& LLM Generation \& Human Filtered & 1 & Execution& \textcolor{red}{\ding{55}}\\
DebugBench~\citep{debugbench}& 3& 18 & Scrape \& LLM Generation \& Human Filtered & 1 & Execution& \textcolor{red}{\ding{55}}\\
CodeEditorBench~\citep{codeeditorbench} & 3& 4& Scrape \& LLM Generation \& Human Filtered & 1 & Execution& \textcolor{red}{\ding{55}}\\
\midrule
\rowcolor{green!15} \benchmark{}~(Ours) & 44& 40 & Online Q\&A& 2 & Human Preference & \textcolor{green}{\ding{51}} \\
\bottomrule
\end{tabular}
}
\caption{Comparison between \benchmark{} and other benchmarks. \benchmark{} provides a comprehensive view by creating diverse user prompts to evaluation alignment between the model-generated response and human preference.}
\label{tab:compare_bench}
\end{table*}
\begin{figure*}[h!]
\centering
\begin{center}
    \includegraphics[width=1.0\textwidth]{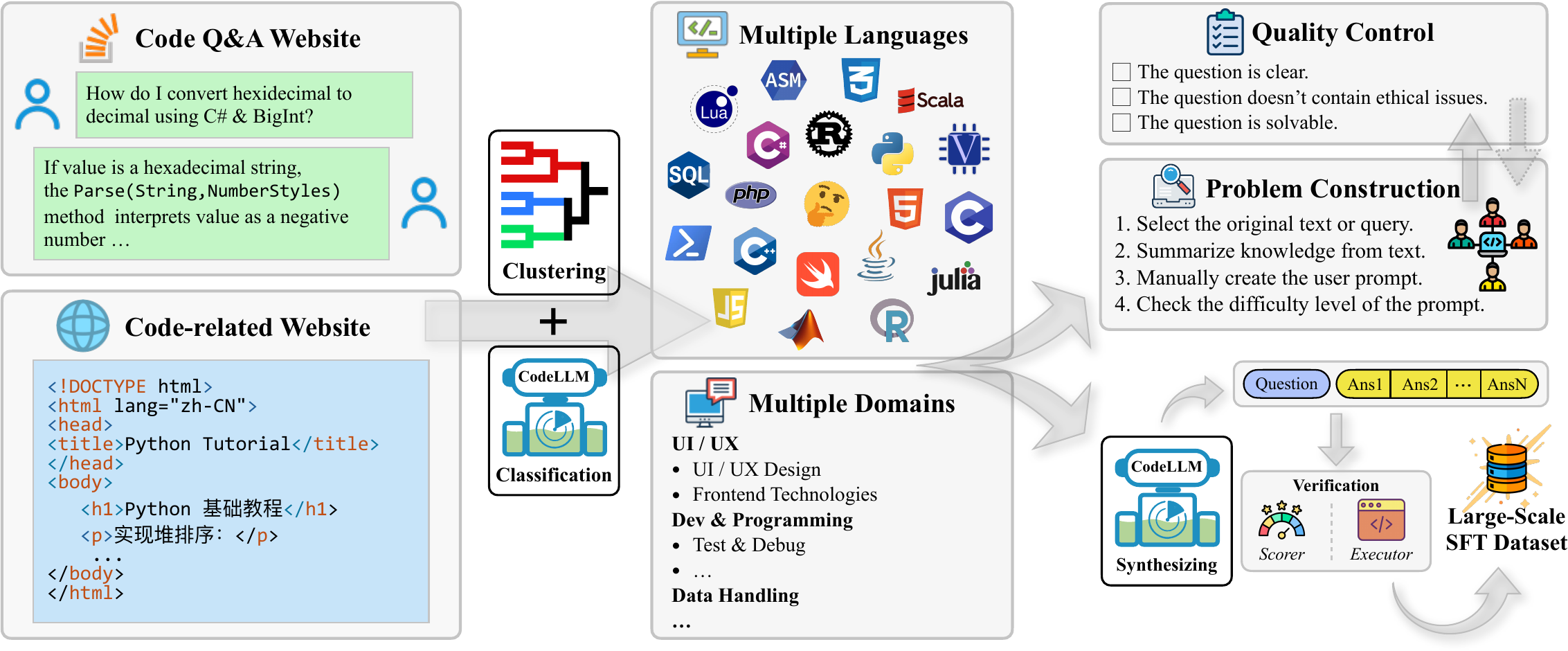}
    \caption{Overview of the \benchmark{} creation benchmark. We first collect the online code Q\&A and code-related raw text from the website. We cluster the code-related data and classify them into different categories using LLM. We uniformly sample the samples from different subtasks as the seed data for manual annotation.}
    \label{fig:bench_construct}
    \vspace{-10pt}
\end{center}
\end{figure*}
\paragraph{Evaluation}
Inspired by the previous work~\cite{chatbot_arena}, we apply \textit{GPT-4o-2024-08-06} as the judger to evaluate the model performance. Specifically, we use two games ``compare A and B '' and ``compare B and A'' (avoid the relative position of A and B affecting the results) to calculate the win rate of A compared to the baseline B.

\paragraph{Decontainmation.}
To avoid data leakage, we apply decontamination to ensure the uniqueness of prompts in \benchmark{}, by removing exact matches (10-gram word overlap) from MultiPL-E~\cite{multiple}, MBPP~\cite{mbpp}, McEval~\cite{codex}, and NaturalCodeBench~\cite{ncb}.

\paragraph{Comparison with other benchmarks}
We compare \benchmark{} with other code benchmarks. Our benchmark provides a valuable comprehensive benchmark for 40 subtasks and 44 programming languages, which satisfies the evaluation in realistic scenarios. \benchmark{} provides many problems for evaluation under realistic scenarios, which are not suitable for verification through unit testing.

\section{\instruct{}}

\paragraph{Recall from Common Crawl.}
A trained fasttext is used to distinguish the code-related text and other common raw text, which is used to recall and clean potential code data and filter out low-quality content using weak model-based classifiers and scorers. Our approach encompasses both file-level and repository-level pertaining to ensure comprehensive coverage. 

\paragraph{Code Classification for Code Snippet.}
We extract the first layer of CodeBERT \cite{code_bert} and fine-tune the tiny classifier on nearly 100 programming languages to build a language identification model. We keep the main language data (e.g. C, Python, and Java) and downsample high-resource language data (e.g. HTML and Java) to keep the balance. Besides, we also remove the samples with no code snippets.

\begin{figure}[t!]
\centering
\includegraphics[width=0.85\columnwidth]{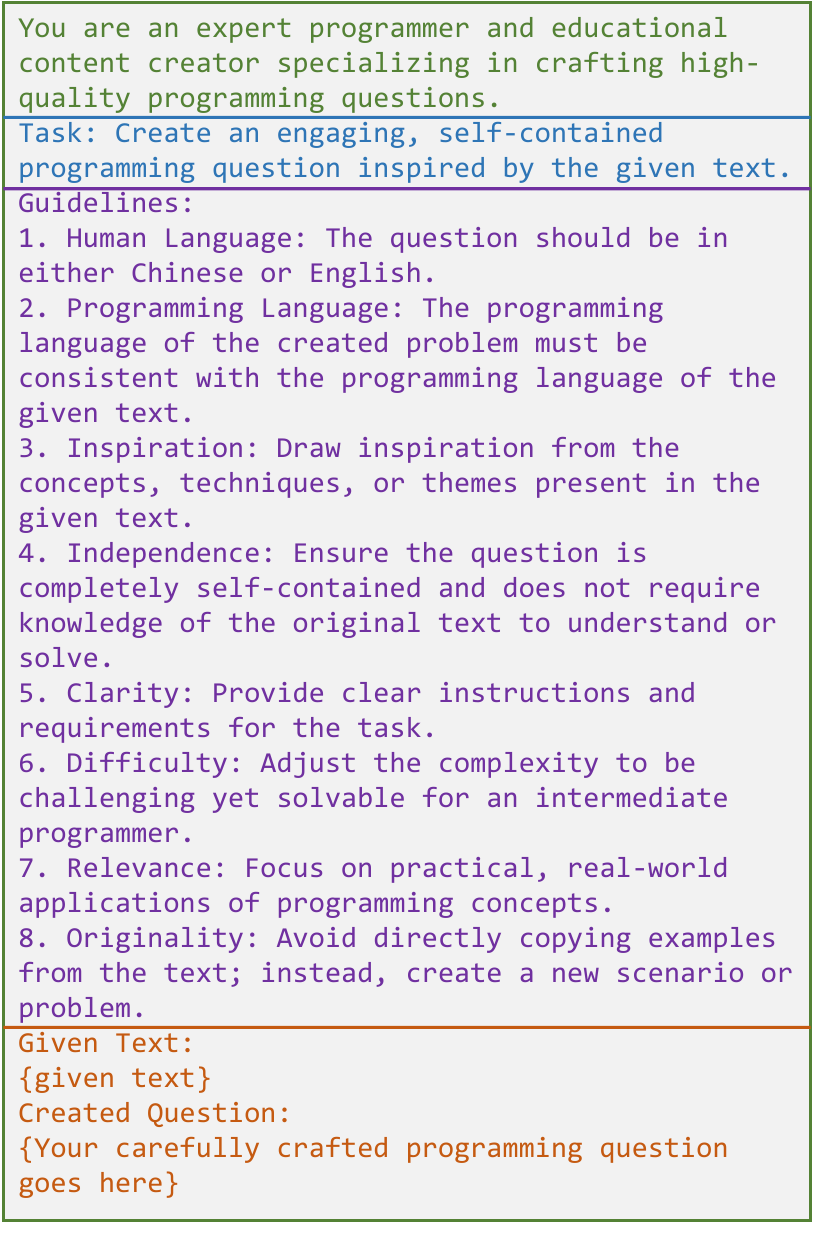}
\caption{Prompt of generating large-scale self-contained synthetic instruction data.}
\vspace{-10pt}
\label{fig:prompt_syncc}
\end{figure}

\paragraph{Scaling Code Instruction}
Initially, we adopt rule-based filtering to clean pre-extracted content from recalled documents by removing site information, advertisements, and HTML tags, thereby significantly reducing document length for further processing. Different from the previous work~\cite{mammoth2}, we utilize Qwen2.5-72B to create new questions instead of extracting question and answer pairs. As shown in Figure \ref{fig:prompt_syncc}. We use the Qwen2.5-Coder to generate multiple responses by sampling for the same document. For the algorithmic generated question and answer, we first adopt a fine-tuned generator to generate the test cases and adopt the multilingual sandbox to verify the correctness of the generated code snippet. As shown in Figure \ref{fig:bench_construct}, for the non-algorithmic query, we first randomly generate four candidates (Best-of-N) and use the LLM to score the candidates (LLM scorer), where the candidates are fed into the LLM to select the best response with the reason. For the algorithmic queries, the generated test cases by LLM are used to verify the correctness of the responses (Executor).
Finally, we select the response with the best score as the response to create \instruct{}. The synthetic instruction corpora generated by Qwen2.5 is used for the first stage and the high-quality data from GPT-4o is used for the second stage.

\begin{table*}[h!]
 \centering
 \resizebox{1.0\textwidth}{!}{
 \begin{tabular}{lr|ccccccc|c}
 \toprule
 \textbf{Model} & \textbf{Size} & \makecell[c]{UI\&UX} & \makecell[c]{Development\& \\ Programming} & \makecell[c]{Specialized \\ Computing} & \makecell[c]{Tools, Environs, \\ \& Practices} & \makecell[c]{Emerging Techs\\\&Apps} & \makecell[c]{Miscellaneous \& \\ General Inquiry} & \makecell[c]{Databases\&\\ Data Handling} & \textbf{Avg.} \\
 \midrule
 \multicolumn{10}{c}{\textbf{Proprietary LLMs and 200B+ LLMs}} \\ \midrule
 Claude-3.5-Sonnet-20240620 & \faLock{} & 88.9/2.2 & 77.3/13.6 & 74.2/18.0 & 81.4/11.9 & 78.9/10.5 & 71.4/28.6 & 63.6/4.5 & 77.8/12.5 \\
 Claude-3.5-Sonnet-20241022 & \faLock{} & 82.2/6.7 & 75.8/12.9 & 76.4/16.9 & 84.7/10.2 & 84.2/13.2 & 57.1/28.6 & 68.2/22.7 & 78.1/13.5 \\
 GPT-3.5-turbo-0125 & \faLock{} & 17.8/24.4 & 11.4/20.5 & 4.5/19.1 & 11.9/18.6 & 10.5/21.1 & 13.6/9.1 & 0.0/14.3 & 10.5/19.6 \\
 GPT-4o-mini-2024-07-18 & \faLock{} & 71.1/13.3 & 62.1/17.4 & 50.0/13.6 & 65.2/14.6 & 72.9/13.6 & 71.1/18.4 & 71.4/14.3 & 65.8/15.6 \\
 GPT-4o-2024-08-06 & \faLock{} & 66.7/17.8 & 72.7/19.7 & 62.9/19.1 & 69.5/15.3 & 76.3/13.2 & 85.7/14.3 & 59.1/22.7 & 69.1/18.1 \\
 o1-mini & \faLock{} & \underline{93.3/4.4} & \underline{94.7/2.6} & 84.1/7.6 & \underline{91.0/5.6} & 88.1/3.4 & \underline{95.5/0.0} & \underline{100.0/0.0} & \underline{89.3/5.1} \\
 o1-preview & \faLock{} & 93.3/2.2 & 81.8/7.6 & \underline{85.4/7.9} & 78.0/6.8 & \underline{92.1/2.6} & 77.3/4.5 & 71.4/28.6 & 83.9/6.6 \\
 Yi-lightning & \faLock{} & 62.2/15.6 & 60.0/11.5 & 57.9/5.3 & 49.4/16.9 & 71.2/11.9 & 54.5/13.6 & 85.7/0.0 & 59.5/12.6 \\
 Doubao-Pro & \faLock{} & 51.1/20.0 & 40.8/18.5 & 55.3/26.3 & 38.2/19.1 & 47.5/22.0 & 36.4/31.8 & 42.9/57.1 & 43.6/21.5 \\
 Qwen-Max & \faLock{} & 75.6/17.8 & 74.2/13.6 & 59.6/24.7 & 78.0/6.8 & 68.4/23.7 & 100.0/0.0 & 81.8/4.5 & 71.9/15.8 \\\midrule
 \multicolumn{10}{c}{\textbf{0.5B+ Open-source LLMs}} \\ \midrule
 Qwen2.5-0.5B-Instruct & 0.5B & \underline{2.2/4.4} & 4.6/4.6 & \underline{5.3/10.5} & 2.2/4.5 & \underline{3.4/5.1} & \underline{4.5/9.1} & 0.0/14.3 & 3.6/5.6 \\
 Qwen2.5-Coder-0.5B-Instruct & 0.5B & 2.2/2.2 & \underline{4.6/6.9} & 2.6/5.3 & \underline{4.5/2.2} & \underline{3.4/5.1} & 4.5/0.0 & \underline{28.6/14.3} & \underline{4.4/4.6} \\
 \midrule
 \multicolumn{10}{c}{\textbf{1B+ Open-source LLMs}} \\ \midrule
 DS-Coder-1.3B-Instruct & 1.3B & \underline{66.7/2.2} & 2.3/5.4 & 2.6/10.5 & 1.7/6.8 & 0.0/9.1 & 2.2/3.4 & 0.0/14.3 & 2.6/5.6 \\
 Yi-Coder-1.5B-Chat & 1.5B & 11.1/2.2 & 5.1/3.4 & 5.4/4.6 & 2.6/5.3 & 2.2/5.6 & 4.5/4.5 & 14.3/14.3 & 7.4/5.1 \\
 Qwen2.5-Coder-1.5B-Instruct & 1.5B & 11.1/4.4 & \underline{15.9/9.1} & \underline{9.0/16.9} & \underline{13.6/11.9} & \underline{13.2/5.3} & \underline{14.3/42.9} & \underline{18.2/4.5} & \underline{13.2/10.7} \\
 OpenCoder-1.5B-Instruct & 1.5B & 11.1/4.4 & 3.8/5.4 & 0.0/5.3 & 2.2/4.5 & 3.4/8.5 & 4.5/9.1 & 0.0/0.0 & 6.7/3.8 \\
 \midrule
 \multicolumn{10}{c}{\textbf{3B+ Open-source LLMs}} \\ \midrule
 Qwen2.5-Coder-3B-Instruct & 3B & \underline{35.6/11.1} & \underline{29.5/10.6} & \underline{27.0/15.7} & \underline{20.3/18.6} & \underline{28.9/10.5} & \underline{42.9/14.3} & \underline{27.3/13.6} & \underline{28.3/13.3} \\
 \midrule
 \multicolumn{10}{c}{\textbf{6B+ Open-source Models}} \\ \midrule
 CodeLlama-7B-Instruct & 7B & 33.3/8.9 & 28.8/18.6 & 23.8/13.8 & 18.2/9.1 & 31.6/5.3 & 29.2/14.6 & \underline{71.4/0.0} & 28.2/12.8 \\
 Llama3-8B-Instruct & 7B & 20.0/17.8 & 14.6/11.5 & 15.8/2.6 & 13.5/9.0 & 16.9/11.9 & 22.7/0.0 & 57.1/14.3 & 16.7/10.3 \\
 Llama3.1-8B-Instruct & 7B & 2.2/8.9 & 4.5/10.1 & 3.8/6.2 & 3.4/6.8 & 5.3/2.6 & 9.1/9.1 & 14.3/0.0 & 7.9/4.4 \\
 DS-Coder-6.7B-Instruct & 6.7B & 11.1/17.8 & 13.1/13.8 & 13.6/8.5 & 13.2/7.9 & 9.0/7.9 & 13.6/4.5 & 28.6/0.0 & 12.3/10.8 \\
 CodeQwen1.5-7B-Chat & 7B & 17.8/15.6 & 13.8/12.3 & 15.8/0.0 & 15.7/9.0 & 15.3/15.3 & 18.2/13.6 & 14.3/42.9 & 15.4/11.8 \\
 Yi-Coder-9B-Chat & 9B & 15.6/17.8 & 15.4/9.2 & 15.8/7.9 & 13.5/13.5 & 10.2/20.3 & 18.2/13.6 & 28.6/28.6 & 14.6/13.3 \\
 DS-Coder-V2-Lite-Instruct & 2.4/16B & \underline{42.2/20.0} & 33.3/17.4 & 31.5/16.9 & 35.6/20.3 & \underline{39.5/21.1} & \underline{71.4/14.3} & 31.8/22.7 & 35.5/18.6 \\
 Qwen2.5-Coder-7B-Instruct & 7B & 40.0/22.2 & \underline{46.2/19.7} & \underline{43.8/15.7} & \underline{40.7/20.3} & 34.2/15.8 & 71.4/0.0 & 40.9/22.7 & \underline{43.1/18.6} \\
 OpenCoder-8B-Instruct & 8B & 24.4/8.9 & 14.6/8.5 & 10.5/7.9 & 9.0/4.5 & 13.6/6.8 & 18.2/9.1 & 14.3/0.0 & 14.1/7.1 \\
 \midrule
 \multicolumn{10}{c}{\textbf{13B+ Models}} \\ \midrule
 CodeLlama-13B-Instruct & 13B & 13.3/4.4 & 7.9/6.7 & 6.8/8.5 & 7.7/6.2 & 4.5/4.5 & 5.3/5.3 & 14.3/14.3 & 11.2/7.9 \\
 Starcoder2-15B-Instruct-v0.1 & 15B & 6.7/6.7 & 6.8/12.9 & 4.5/15.7 & 6.8/6.8 & 5.3/13.2 & 13.6/13.6 & 0.0/14.3 & 6.4/12.0 \\
 Qwen2.5-Coder-14B-Instruct & 14B & \underline{51.1/24.4} & \underline{53.0/17.4} & \underline{52.8/16.9} & \underline{50.8/18.6} & \underline{57.9/7.9} & \underline{28.6/28.6} & \underline{36.4/27.3} & \underline{60.6/51.5} \\
 \midrule
 \multicolumn{10}{c}{\textbf{20B+ Models}} \\ \midrule
 CodeLlama-34B-Instruct & 34B & 11.1/6.7 & 2.6/2.6 & 6.9/2.3 & 8.5/6.8 & 7.9/10.1 & 9.1/9.1 & 14.3/0.0 & 7.7/5.6 \\
 CodeStral-22B-v0.1 & 22B & 17.8/22.2 & 27.3/13.6 & 14.6/14.6 & 25.4/10.2 & 18.4/10.5 & 14.3/42.9 & 22.7/22.7 & 21.7/15.8 \\
 DS-Coder-33B-Instruct & 33B & 13.3/11.1 & 22.0/9.8 & 12.4/12.4 & 13.6/6.8 & 13.2/18.4 & 28.6/42.9 & 22.7/18.2 & 16.8/12.0 \\
 CodeLlama-70B-Instruct & 70B & 11.1/22.2 & 9.2/10.0 & 10.5/5.3 & 9.0/6.7 & 16.9/8.5 & 9.1/13.6 & 0.0/0.0 & 15.5/10.5 \\
 DS-Coder-V2-Instruct & 21/236B & 55.6/11.1 & 62.1/18.2 & 60.7/14.6 & 50.8/18.6 & 52.6/21.1 & 71.4/14.3 & 40.9/31.8 & 57.4/17.6 \\
 DS-V2.5 & 21/236B & 77.8/11.1 & \underline{72.0/12.9} & 71.9/13.5 & 71.2/8.5 & 73.7/10.5 & 100.0/0.0 & 68.2/13.6 & 73.0/11.7 \\
 Llama3-70B-Instruct & 7B & 35.6/20.0 & 26.2/26.2 & 25.4/22.0 & 34.2/15.8 & 23.6/14.6 & 36.4/4.5 & 14.3/57.1 & 27.7/20.5 \\
 Llama3.1-70B-Instruct & 7B & 48.9/24.4 & 43.8/20.0 & 34.2/26.3 & 40.4/22.5 & 54.2/20.3 & 45.5/9.1 & 71.4/14.3 & 44.9/21.0 \\
 Qwen2.5-Coder-32B-Instruct & 32B & 71.1/13.3 & 66.7/15.9 & 67.4/16.9 & 74.6/13.6 & 65.8/18.4 & \underline{100.0/0.0} & 63.6/18.2 & 68.9/15.6 \\
 Qwen2.5-32B-Instruct & 32B & 62.2/15.6 & 52.3/15.4 & 57.9/18.4 & 50.6/23.6 & 54.2/13.6 & 50.0/13.6 & 71.4/14.3 & 54.1/17.1 \\
 QwQ-32B-Preview & 32B & 53.3/15.6 & 56.8/16.7 & 50.6/16.9 & 64.4/5.1 & 52.6/21.1 & 85.7/0.0 & 63.6/9.1 & 56.6/14.5 \\
 Qwen2.5-72B-Instruct & 72B & \underline{82.2/6.7} & 71.5/14.6 & \underline{76.3/13.2} & \underline{75.3/15.7} & \underline{71.2/18.6} & 63.6/13.6 & \underline{85.7/14.3} & \underline{73.8/14.4} \\
 \rowcolor{green!15} \baseline{} & 32B & 55.6/26.7 & 49.2/20.8 & 36.8/36.8 & 50.6/20.2 & 52.5/20.3 & 40.9/18.2 & 57.1/0.0 & 49.2/22.3 \\
 \bottomrule
 \end{tabular}
 }
 \caption{The win/tie rate of different instruction LLMs on \benchmark{}. The underlined numbers represent the best scores within the same model size range.}
 \label{tab:CodeArena}
 \vspace{-10pt}
\end{table*}

\section{Experimental Setup}
\label{sec:experimental_setup}

\subsection{Instruction Dataset}

\paragraph{CodeLLMs}
We evaluate 40+ models with sizes ranging from 0.5B to 200B parameters, including general/code LLMs and open/closed-source models. For general models, we evaluate GPTs~\citep{gpt3,gpt4} (GPT-3.5-Turbo, GPT4-o), Qwen series (Qwen2.5 and Qwen-Max)~\citep{Qwen}, Claude series~\cite{claude}, Llama3/3.1~\citep{llama3}, Yi~\citep{yi}, and o1 series. For code models, we test CodeLlama~\cite{code_llama}, OpenCoder~\cite{opencoder}, Qwen-Coder~\citep{qwen25coder}, DeepSeekCoder~\citep{deepseek_coder}, and CodeStral~\citep{codestral}.

\subsection{Evaluation Benchmark}

\paragraph{EvalPlus and MultiPL-E.} 
The EvalPlus~\citep{evalplus} is an upgraded version of the HumanEval~\citep{codex} and MBPP~\citep{mbpp} to test the code generation capabilities. The benchmark reports the scores of HumanEval (HE)/MBPP with base test cases and HumanEval+ (HE+)/MBPP+ with plus test cases.
\paragraph{MultiPL-E} The MultiPL-E test set~\cite{multiple} contains the HumanEval (Python) and translated test set of other programming languages, i.e., Java, C++, Javascript, and Typescript.
\paragraph{CodeArena}
Different from the EvalPlus and MultiPL-E, \benchmark{} consists of many non-algorihtmic, which is not suitable for code-execution-based evaluation. Each question is scored twice to calculate the win rate and tie rate by GPT-4o using a different input order ``A, B'' and ``B, A'', where ``A'' is the baseline from \texttt{gpt-4-turbo-2024-04-09} and ``B'' is the model-generated response.

\subsection{Evaluation Metrics}
\paragraph{Pass@k}
Given the model-generated response, we extract the expected function and feed the test cases into the extracted function to verify the correctness of the generation. We adopt greedy Pass@1~\cite{codex} to report the results on EvalPlus and MultiPL-E,
\paragraph{LLM as a judgement} 
Due to the high cost of collecting human preferences~\cite{llm_as_a_judge}, we use pairwise comparison for judgment, where an LLM judger is fed with a question and two answers and determines which one is better or declares a tie\footnote{\url{https://github.com/lmarena/arena-hard-auto}}. We report win rate/tie rate for \benchmark{}.

\subsection{Impletmentation Details}
We fine-tune Qwen2.5-Coder-32B on nearly 20B synthetic tokens generated from website data, where GPT-4o generates 1B tokens and Qwen2.5-Coder-Instruct generates the left tokens. \baseline{} is fine-tuned on the synthetic instruction corpus \instruct{} with $256$ NVIDIA A100-80GB GPUs. The learning rate first increases into $3 \times 10^{-4}$ with $100$ warmup steps and then adopts a cosine decay scheduler. We adopt the Adam optimizer~\cite{adam} with a global batch size of $2048$ samples and a tensor parallel size of 8, truncating sentences to 32K tokens.

\section{Results and Discussion}
\label{sec:results_and_discussion}

\begin{table*}[h!]
 \centering
 \resizebox{0.85\textwidth}{!}{
 \begin{tabular}{lr|cccc|cccccccc|c}
 \toprule
 \textbf{Model} & \textbf{Size} & HE & HE+ & MBPP & MBPP+ & Python & Java & C++ & C\# & TS & JS & PHP & Bash & \textbf{Avg.} \\
 \midrule
 \multicolumn{15}{c}{\textbf{Closed-APIs}} \\ \midrule
 Claude-3.5-Sonnet-20240620 & \faLock{} & 89.0 & 81.1 & 87.6 & 72.0 & 89.6 & 86.1 & 82.6 & 85.4 & 84.3 & 84.5 & 80.7 & 48.1 & 80.2 \\
 Claude-3.5-Sonnet-20241022 & \faLock{} & 92.1 & 86.0 & 91.0 & 74.6 & 93.9 & 86.7 & 88.2 & \underline{87.3} & 88.1 & 91.3 & 82.6 & 52.5 & 83.8 \\
 GPT-4o-mini-2024-07-18 & \faLock{} & 87.8 & 84.8 & 86.0 & 72.2 & 87.2 & 75.9 & 77.6 & 79.7 & 79.2 & 81.4 & 75.2 & 43.7 & 75.0 \\
 GPT-4o-2024-08-06 & \faLock{} & 92.1 & 86.0 & 86.8 & 72.5 & 90.9 & 83.5 & 76.4 & 81.0 & 83.6 & 90.1 & 78.9 & 48.1 & 79.1 \\
 o1-mini & \faLock{} & \underline{97.6} & \underline{90.2} & \underline{93.9} & \underline{78.3} & 95.7 & \underline{90.5} & \underline{93.8} & 77.2 & \underline{91.2} & 92.5 & 84.5 & \underline{55.1} & 85.1 \\
 o1-preview & \faLock{} & 95.1 & 88.4 & 93.4 & 77.8 & \underline{96.3} & 88.0 & 91.9 & 84.2 & 90.6 & \underline{93.8} & \underline{90.1} & 47.5 & \underline{85.3} \\
 \midrule
 \multicolumn{15}{c}{\textbf{0.5B+ Models}} \\ \midrule
 Qwen2.5-Coder-0.5B-Instruct & 0.5B & \underline{61.6} & \underline{57.3} & 52.4 & \underline{43.7} & \underline{61.6} & \underline{57.3} & \underline{52.4} & \underline{43.7} & \underline{50.3} & \underline{50.3} & \underline{52.8} & \underline{27.8} & \underline{49.6} \\
 \midrule
 \multicolumn{11}{c}{\textbf{1B+ Models}} \\ \midrule
 DS-Coder-1.3B-Instruct & 1.3B & 65.9 & 60.4 & 65.3 & 54.8 & 65.2 & 51.9 & 45.3 & 55.1 & 59.7 & 52.2 & 45.3 & 12.7 & 48.4 \\
 Yi-Coder-1.5B-Chat & 1.5B & 69.5 & 64.0 & 65.9 & 57.7 & 67.7 & 51.9 & 49.1 & 57.6 & 57.9 & 59.6 & \underline{52.2} & 19.0 & 51.9 \\
 Qwen2.5-Coder-1.5B-Instruct & 1.5B & \underline{70.7} & \underline{66.5} & \underline{69.2} & \underline{59.4} & \underline{71.2} & \underline{55.7} & \underline{50.9} & \underline{64.6} & \underline{61.0} & \underline{62.1} & 59.0 & \underline{29.1} & \underline{56.7} \\
 \midrule
 \multicolumn{15}{c}{\textbf{3B+ Models}} \\ \midrule
 Qwen2.5-Coder-3B-Instruct & 3B & \underline{84.1} & \underline{80.5} & \underline{73.6} & \underline{62.4} & \underline{83.5} & \underline{74.7} & \underline{68.3} & \underline{78.5} & \underline{79.9} & \underline{75.2} & \underline{73.3} & \underline{43.0} & \underline{72.1} \\
 \midrule
 \multicolumn{15}{c}{\textbf{6B+ Models}} \\ \midrule
 CodeLlama-7B-Instruct & 7B & 40.9 & 33.5 & 54.0 & 44.4 & 34.8 & 30.4 & 31.1 & 21.6 & 32.7 & - & 28.6 & 10.1 & - \\
 DS-Coder-6.7B-Instruct & 6.7B & 74.4 & 71.3 & 74.9 & 65.6 & 78.6 & 68.4 & 63.4 & 72.8 & 67.2 & 72.7 & 68.9 & 36.7 & 66.1 \\
 CodeQwen1.5-7B-Chat & 7B & 83.5 & 78.7 & 77.7 & 67.2 & 84.1 & 73.4 & 74.5 & 77.8 & 71.7 & 75.2 & 70.8 & 39.2 & 70.8 \\
 Yi-Coder-9B-Chat & 9B & 82.3 & 74.4 & 82.0 & 69.0 & 85.4 & 76.0 & 67.7 & 76.6 & 72.3 & 78.9 & 72.1 & 45.6 & 71.8 \\
 DS-Coder-V2-Lite-Instruct & 2.4/16B & 81.1 & 75.6 & 82.8 & 70.4 & 81.1 & \underline{76.6} & \underline{75.8} & 76.6 & 80.5 & 77.6 & 74.5 & 43.0 & 73.2 \\
 Qwen2.5-Coder-7B-Instruct & 7B & \underline{88.4} & \underline{84.1} & \underline{83.5} & \underline{71.7} & \underline{87.8} & 76.5 & 75.6 & \underline{80.3} & \underline{81.8} & \underline{83.2} & \underline{78.3} & \underline{48.7} & \underline{76.5} \\
 OpenCoder-8B-Instruct & 8B & 83.5 & 78.7 & 79.1 & 69.0 & 83.5 & 72.2 & 61.5 & 75.9 & 78.0 & 79.5 & 73.3 & 44.3 & 71.0 \\
 \midrule
 \multicolumn{15}{c}{\textbf{13B+ Models}} \\ \midrule
 CodeLlama-13B-Instruct & 13B & 40.2 & 32.3 & 60.3 & 51.1 & 42.7 & 40.5 & 42.2 & 24.0 & 39.0 & - & 32.3 & 13.9 & - \\
 Starcoder2-15B-Instruct-v0.1 & 15B & 67.7 & 60.4 & 78.0 & 65.1 & 68.9 & 53.8 & 50.9 & 62.7 & 57.9 & 59.6 & 53.4 & 24.7 & 54.0 \\
 Qwen2.5-Coder-14B-Instruct & 14B & \underline{89.6} & \underline{87.2} & \underline{86.2} & \underline{72.8} & \underline{89.0} & \underline{79.7} & \underline{85.1} & \underline{84.2} & \underline{86.8} & \underline{84.5} & \underline{80.1} & \underline{47.5} & \underline{79.6} \\
 \midrule
 \multicolumn{15}{c}{\textbf{20B+ Models}} \\ \midrule
 CodeLlama-34B-Instruct & 34B & 48.2 & 40.2 & 61.1 & 50.5 & 41.5 & 43.7 & 45.3 & 31.0 & 40.3 & - & 36.6 & 19.6 & - \\
 CodeStral-22B-v0.1 & 22B & 81.1 & 73.2 & 78.2 & 62.2 & 81.1 & 63.3 & 65.2 & 43.7 & 68.6 & - & 68.9 & 42.4 & - \\
 DS-Coder-33B-Instruct & 33B & 81.1 & 75.0 & 80.4 & 70.1 & 79.3 & 73.4 & 68.9 & 74.1 & 67.9 & 73.9 & 72.7 & 43.0 & 69.2 \\
 CodeLlama-70B-Instruct & 70B & 72.0 & 65.9 & 77.8 & 64.6 & 67.8 & 58.2 & 53.4 & 36.7 & 39.0 & - & 58.4 & 29.7 & - \\
 DS-Coder-V2-Instruct & 21/236B & 85.4 & 82.3 & 89.4 & 75.1 & 90.2 & \underline{82.3} & \underline{84.8} & 82.3 & 83.0 & 84.5 & \underline{79.5} & \underline{52.5} & \underline{79.9} \\
 Qwen2.5-Coder-32B-Instruct & 32B & \underline{92.7} & 87.2 & 90.2 & 75.1 & \underline{92.7} & 80.4 & 79.5 & \underline{82.9} & \underline{86.8} & \underline{85.7} & 78.9 & 48.1 & 79.4 \\
 Qwen2.5-32B-Instruct & 32B & 87.8 & 82.9 & 86.8 & 70.9 & 88.4 & 80.4 & 81.0 & 74.5 & 83.5 & 82.4 & 78.3 & 46.8 & 76.9 \\
 Qwen2.5-72B-Instruct & 32B & 85.4 & 79.3 & \underline{90.5} & \underline{77.0} & 82.9 & 81.0 & 80.7 & 81.6 & 81.1 & 82.0 & 77.0 & 48.7 & 75.1 \\
 \rowcolor{green!15} \baseline{} & 32B & \underline{92.7} & \underline{87.8} & 86.2 & 74.7 & 92.1 & 80.4 & 80.7 & 81.6 & 83.0 & \underline{85.7} & 77.6 & 49.4 & 78.8 \\
 \bottomrule
 \end{tabular}
 }
 \caption{The performance of different instruction LLMs on EvalPlus and MultiPL-E. ``HE'' denotes the HumanEval, ``HE+'' denotes the plus version with more test cases, and ``MBPP+'' denotes the plus version with more test cases.}
 \label{tab:evalplus_multiple}
\end{table*}

\subsection{Main Results}
\label{subsec:results}

\paragraph{CodeArena.}
Table \ref{tab:CodeArena} shows that the win rate/tie rate of different instruction LLM on \benchmark{}. The closed-source LLMs such as Claude and o1 series still get a dominant advantage compared to Qwen2.5-Coder and DeepseekCoder. There still exists a notable performance gap between open codeLLMs (e.g. Qwen-Coder) and closed-source LLMs (e.g., o1 and Claude series), emphasizing the importance of alignment between model-generated response human preference. \baseline{} totally trained on the large-scale synthetic instruction corpus \instruct{} can still get a strong performance on \benchmark{}, which verifies the correctness of the route of taking large-scale synthetic data to improve model performance.

\paragraph{EvalPlus and MultiPL-E.}
Table \ref{tab:evalplus_multiple} shows that \baseline{} significantly beats previous strong open-source baselines using large-scale synthetic instruction, closing the gap with GPT-4o and Claude, which verifies that the large-scale synthetic data can bring more significant improvement for the base model in the code-execution-based benchmark (code generation) compared to \benchmark{}.

\subsection{Discussion}
\label{subsec:discussion}
\begin{figure*}[t!]
\begin{center}
    \includegraphics[width=1.0\textwidth]{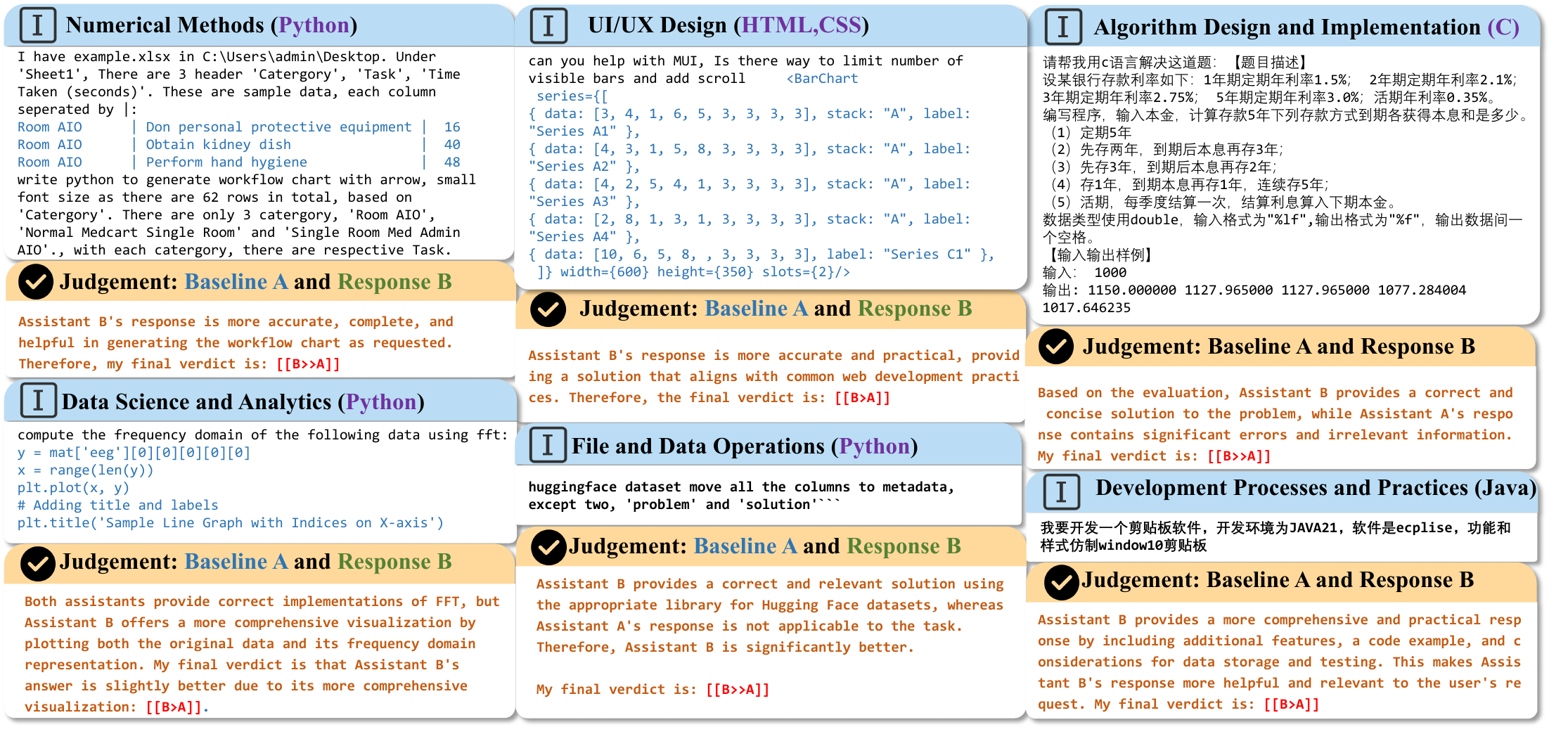}
    \caption{Examples of CodeArena. The LLM judger decides which response is better.}
    \label{fig:examples}
    \vspace{-10pt}
\end{center}
\end{figure*}
\paragraph{Examples of \benchmark{}.}
Figure \ref{fig:examples} lists six examples from the different subtasks, covering Python, HTML, CSS, and Java. Different from the previous benchmarks~\cite{multiple,livecodebench} comprised of algorithmic questions in a fixed format, the queries of \benchmark{} are more consistent with the distribution of user questions in real Q\&A scenarios. For example, the query \textcolor{blue!50}{``huggingface dataset move all the columns to metadata,
except two, problem and solution''} is closer to the question style of real users. For the baseline response A and model-generated response B, the GPT4o thinks B beats A based on the judgment ``B provides a correct and relevant solution using the appropriate library for Hugging Face datasets'', which select responses that are more aligned with human preferences.

\paragraph{Difference between \benchmark{} and Execution-based Benchmark.}
\begin{figure}[h!]
\centering
\includegraphics[width=1.0\columnwidth]{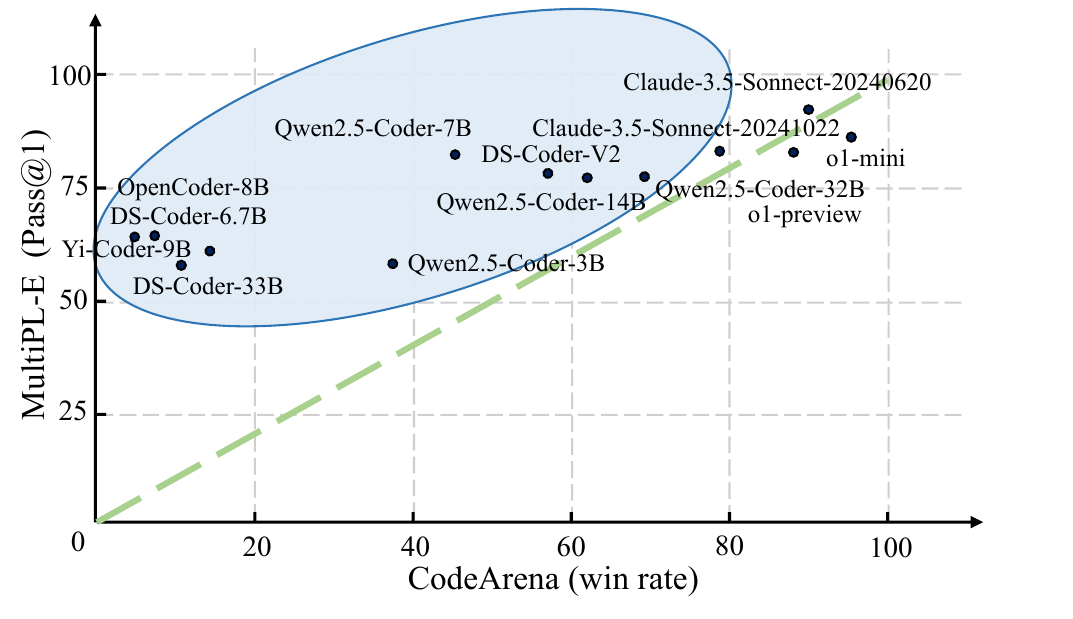}
\caption{Comparison between MultiPL-E and \benchmark{}. LLMs in the blue circle present relatively mismatched performances on two benchmarks. }
\vspace{-5pt}
\label{fig:difference}
\end{figure}
Compared to the benchmark MultiPL-E evaluated by code execution, \benchmark{} is created from real-world Q\&A and evaluated by LLM-as-a-judge to evaluate the alignment between the model-generated response and human preference. For example, the LLMs tend to only generate the code without any natural description (even the code is correct) will bring an unsatisfactory experience to users, which will also lead to poor performance in \benchmark{}.
In Figure \ref{fig:difference}, we can observe that the state-of-the-art closed-source LLMs (e.g. o1 and Claude series) get a balanced performance between the code execution benchmark and \benchmark{}. The open-source models (e.g. DeepseekCoder and Qwen-Coder) are likely to bring a bad experience to users, where the generated response lacks a more detailed explanation or more complete details compared to closed-source LLMs.

\paragraph{Scaling Synthetic Instruction Corpora.}
\begin{figure}[h!]
\centering
\includegraphics[width=0.85\columnwidth]{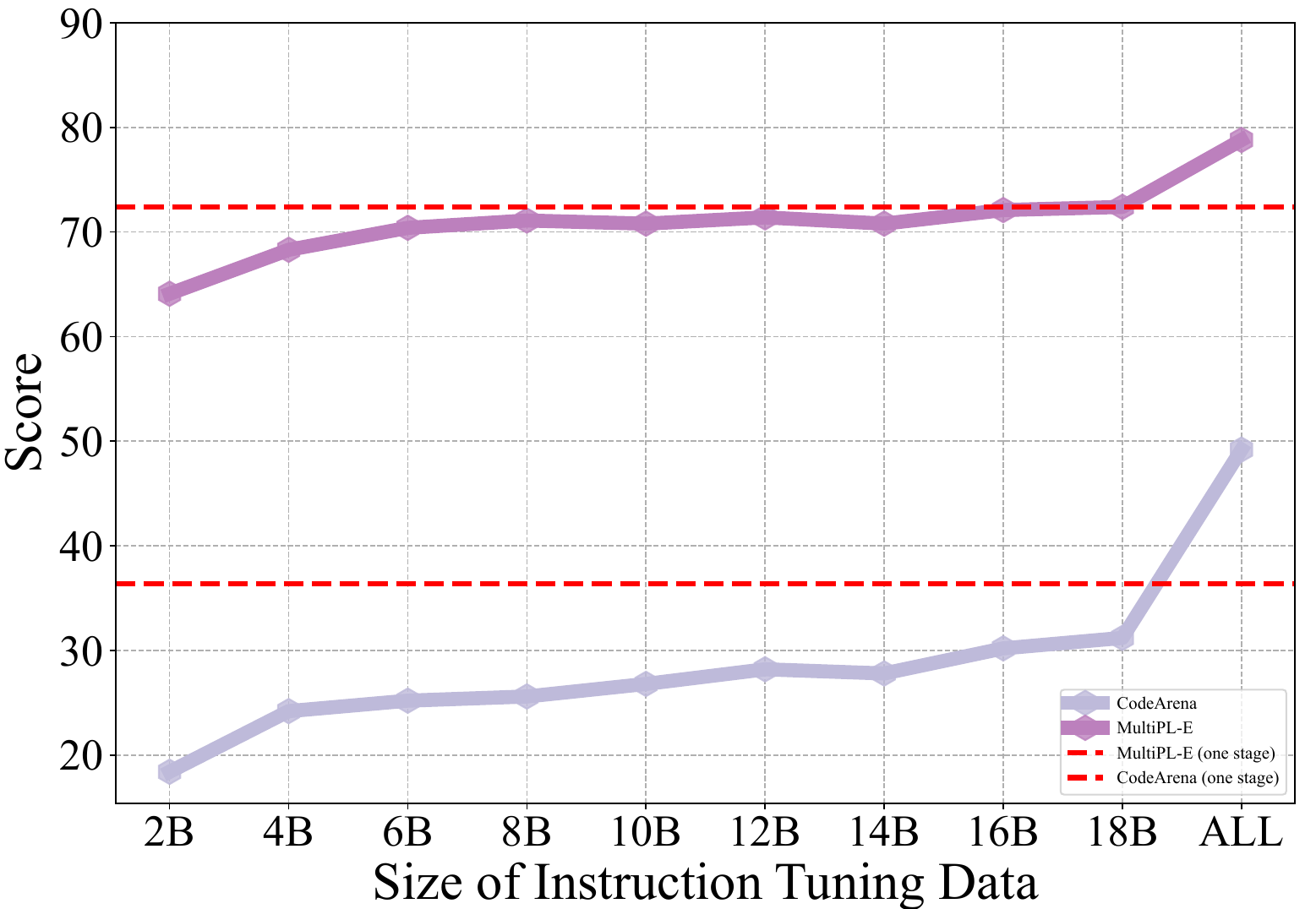}
\caption{Results of \benchmark{} with different data size on MultiPL-E and \benchmark{}.}
\vspace{-5pt}
\label{fig:scaling_code_arena}
\end{figure}
We would like to further analyze the performance of \baseline{} in MultiPl-E and \benchmark{} given different sizes of instruction corpora. Therefore, we select the full instruction (19B synthetic data is at the front of the data and 1B high-quality data is at the end) set \instruct{} and extract the first $K$ billion tokens as the fine-tuned data. We set $K=\{2,4,\dots,20\}$. We randomly extract specific data from the whole sentence pairs.
Figure~\ref{fig:scaling_code_arena} shows the performance on \benchmark{}. With the increase of instruction data, \baseline{} still can get significant improvement, which emphasizes the importance of the scaling instruction corpora. Besides, the two-stage SFT gets a better performance compared to the one-stage training (red line), where the high-quality data brings a huge improvement at last.

\paragraph{Distribution of different benchmarks.}
\begin{figure}[h!]
\centering
\includegraphics[width=0.85\columnwidth]{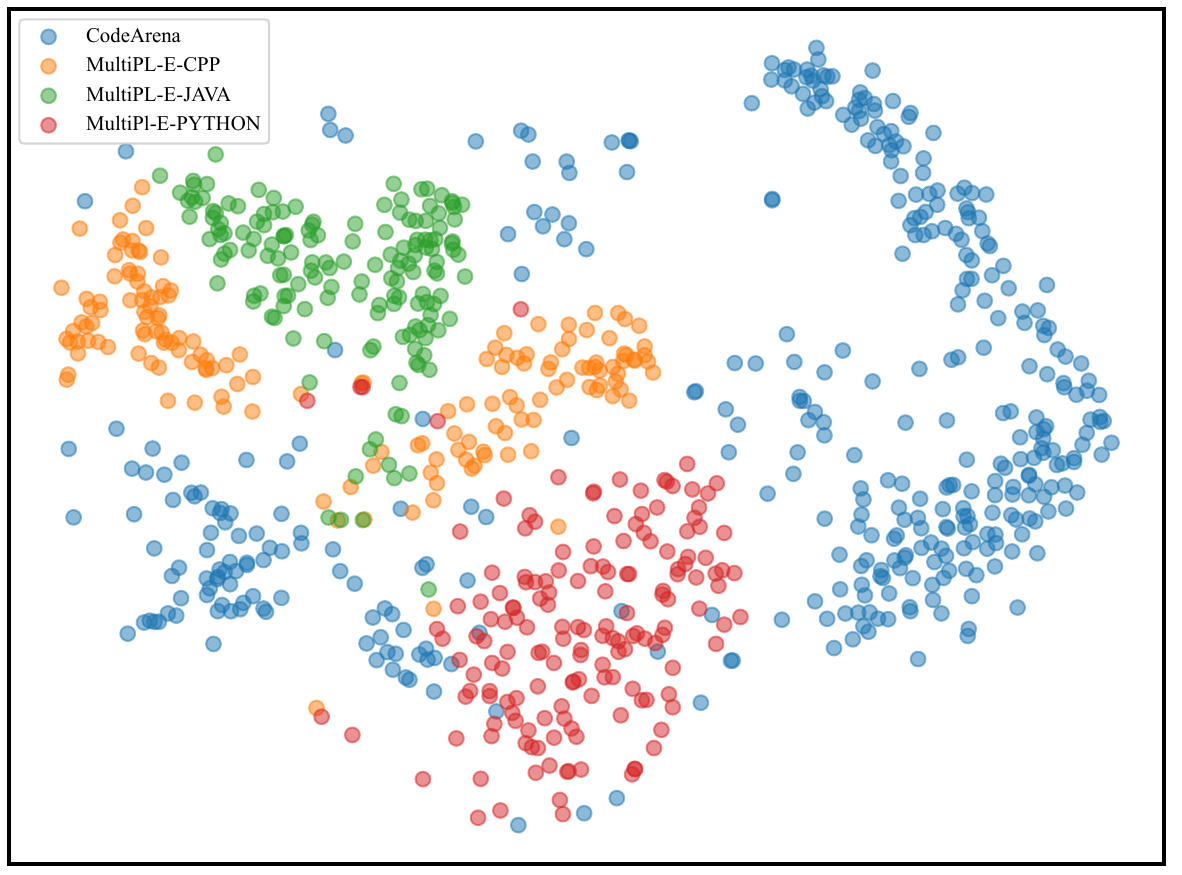}
\caption{Distribution of \benchmark{} and MultiPL-E of different languages.}
\vspace{-5pt}
\label{fig:benchmark_distribution}
\end{figure}
We visualize the queries of \benchmark{} and MultiPL-E (Python, Java, and CPP) by extracting the encoder representations of the last layer for t-SNE~\cite{tsne}. The average of all hidden states of the last encoder layer is regarded as the query representation. In Figure \ref{fig:benchmark_distribution}, the representations of \benchmark{} are distributed in the whole area, while the representations of different languages in MultiPL-E are separately located in a narrow area. It shows that the distribution of queries in \benchmark{} is very diverse, which is suitable for evaluating human preferences in realistic scenarios.

\section{Related Work}
\label{sec:related_work}
\paragraph{Code Large Language Model.}
Large language models (LLMs) designed for coding tasks have demonstrated exceptional capabilities in code generation, debugging, translation, and other essential functions for modern software engineering~\citep{Chen2021Evaluating,claude,gpt4,fried2022incoder,xu2022systematic,unicoder}. Numerous in-file benchmarks have been developed to evaluate these capabilities; however, many of them focus on a limited selection of programming languages, such as Python and Java~\citep{codegeex,mbpp,livecodebench}. Recent advancements in code LLMs, including models like Code Llama~\citep{codellama}, DeepSeek-Coder~\citep{deepseek_coder}, OpenCoder~\citep{opencoder}, and Qwen2.5-Coder~\citep{qwen25coder}, have made significant strides in multilingual code generation and debugging tasks. These models have been effectively evaluated using benchmarks such as MultiPL-E~\citep{multiple}, McEval~\citep{mceval}, and MdEval~\citep{mdevl}.

\paragraph{Code Benchmarks.}
Code generation is a basic task for code language models (LLMs), requiring them to interpret natural language descriptions and generate corresponding code snippets that fulfill user requirements~\citep{cruxeval,ds1000,evalplus,yu2024codereval,autokaggle}. To thoroughly evaluate the diverse capabilities of LLMs, numerous benchmarks have been proposed, including code translation~\citep{jiao2023evaluation,codetransocean,zhu2022xlcost}, code retrieval~\citep{huang2021cosqa,codesearchnet,codexglue}, code completion~\cite{fim,liu2024m2rc,repocoder}, code debugging~\citep{review4repair,debugbench,mdevl}, and structured data understanding~\cite{tablebench,tablegpt2}. Recent initiatives such as McEval~\citep{mceval} have expanded the evaluative scope to 40 programming languages for multilingual scenarios, while MdEval~\citep{mdevl} has developed a multilingual code debugging benchmark encompassing nearly 20 programming languages. Nonetheless, many of these studies concentrate on assessing only a single aspect of LLM capabilities, often overlooking the evaluation of LLMs as comprehensive program developers across a variety of real-world coding scenarios. In this work, we propose FullStack Bench to evaluate the capabilities of LLMs across multiple practical code development contexts.

\section{Conclusion}
\label{sec:conclusion}
In this work, We introduce \benchmark{}, a meticulously human-curated benchmark composed of 397 high-quality samples spanning 40 categories, derived from real-world user queries, to address discrepancies between model-generated responses and human preferences in coding tasks. Additionally, we create \instruct{}, a diverse synthetic instruction corpus containing nearly 20 billion tokens, by scaling web-sourced instructions. Our evaluation of over 20 large language models (LLMs) using \benchmark{} highlights significant performance discrepancies between code-execution-based benchmarks and our human-curated benchmark. Notably, there is a marked performance gap between open-source code LLMs (such as DeepSeek-Coder) and closed-source LLMs (such as the o1 and Claude series), underscoring the importance of aligning AI models with human preferences in coding tasks.




\bibliography{custom}
\bibliographystyle{acl_natbib}


\end{document}